\documentclass[letterpaper, 10 pt, conference]{ieeeconf}

\usepackage{graphicx}
\usepackage{balance}
\usepackage{comment}
\usepackage{cite}
\usepackage{amssymb}
\usepackage[tight,footnotesize]{subfigure}
\usepackage[active]{srcltx}
\usepackage{amsmath}

\graphicspath{{./figs/}}
\renewcommand{\baselinestretch}{0.95}
\usepackage{eurosym}
\usepackage[plain]{algorithm}
\usepackage{algorithmic}
\usepackage{multicol}
\usepackage{dsfont}
\usepackage{amsfonts}

\usepackage{cases}
\usepackage{xcolor}
\IEEEoverridecommandlockouts
\begin{document}

\title{3D Trajectory Planning for UAV-based Search Missions:\\
An Integrated Assessment and Search Planning Approach
}

\author{Savvas~Papaioannou,~Panayiotis~Kolios,~Theocharis~Theocharides,\\~Christos~G.~Panayiotou~ and ~Marios~M.~Polycarpou
\thanks{The authors are with the KIOS Research and Innovation Centre of Excellence (KIOS CoE) and the Department of Electrical and Computer Engineering, University of Cyprus, Nicosia, 1678, Cyprus. {\tt\small \{papaioannou.savvas, pkolios, ttheocharides, christosp, mpolycar\}@ucy.ac.cy}}
}

\maketitle

\begin{abstract}

The ability to efficiently plan and execute search missions in challenging and complex environments during natural and man-made disasters is imperative.  In many emergency situations, precise navigation between obstacles and time-efficient searching around 3D structures is essential for finding survivors. 
In this work we propose an integrated assessment and search planning approach which allows an autonomous UAV (unmanned aerial vehicle) agent to plan and execute collision-free search trajectories in 3D environments. More specifically, the proposed search-planning framework aims to integrate and automate the first two phases (i.e., the assessment phase and the search phase) of a traditional search-and-rescue (SAR) mission. In the first stage, termed assessment-planning we aim to find a high-level assessment plan which the UAV agent can execute in order to visit a set of points of interest. The generated plan of this stage guides the UAV to fly over the objects of interest thus providing a first assessment of the situation at hand. In the second stage, termed search-planning, the UAV trajectory is further fine-tuned to allow the UAV to search in 3D (i.e., across all faces) the objects of interest for survivors. The performance of the proposed approach is demonstrated through extensive simulation analysis.    
\end{abstract}

\section{Introduction} \label{sec:Introduction}

\begin{figure}
	\centering
	\includegraphics[width=\columnwidth]{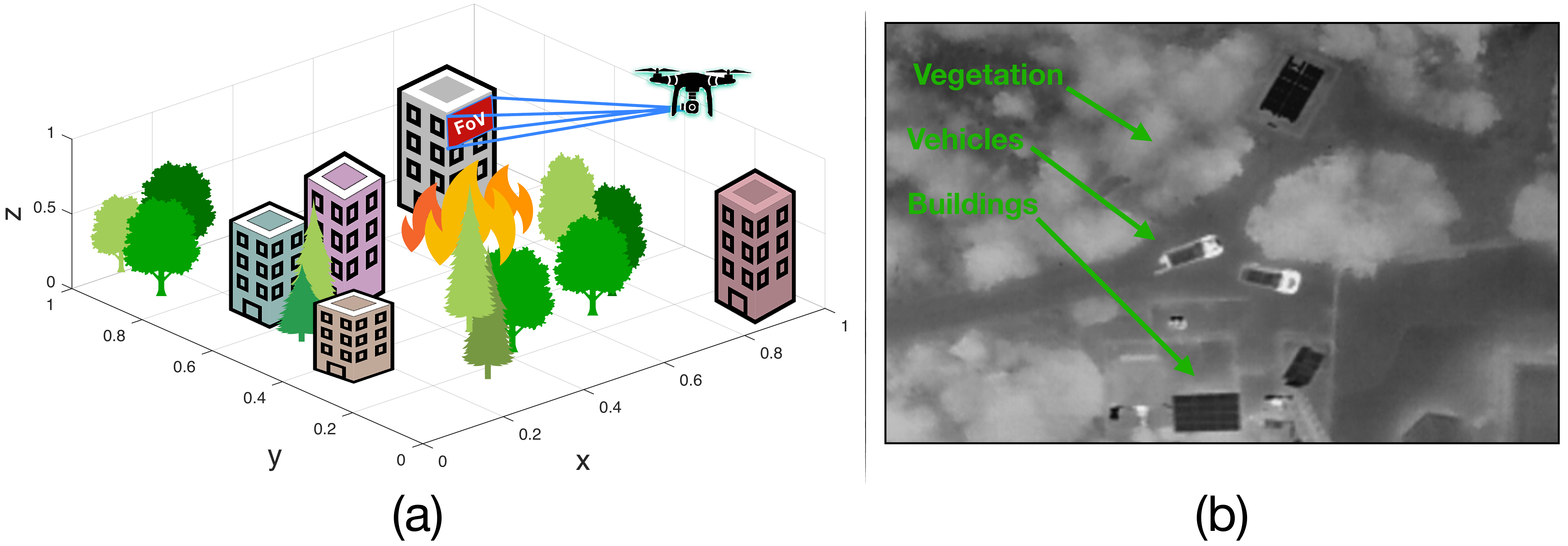}
	\caption{(a) In many emergency situations, precise navigation between obstacles and efficient searching around 3D structures is essential for finding survivors, (b) Aerial view of a rural area taken during a SAR missing person incident from a UAV equipped with a thermal camera.}	
	\label{fig:scenario}
	\vspace{-0mm}
\end{figure}

Over the last several years we have witnessed an unprecedented interest in unmanned aerial systems. Indeed, the miniaturization and cost reduction of electronic components and the recent technological advancements in avionics, robotic systems and artificial intelligence has led to the rapid growth of unmanned aerial vehicles (UAVs). This has enabled the utilization of UAVs in search-and-rescue (SAR) missions \cite{Goodrich2008,Bitton2008,Hayat2017,Bernard2011} i.e., during natural or man-made disasters and in emergency response situations. UAVs are currently being used by first responders mainly to provide a birds-eye view of the incident and for conducting rapid spot searches over inaccessible areas to locate missing people and for damage assessment. However, in many situations searching the affected area for survivors with a birds-eye view in not sufficient, especially in challenging and complex environments with obstacles and occlusions, as illustrated in Fig. \ref{fig:scenario}. In such emergency situations, precise navigation between obstacles and efficient searching around 3D structures is essential for finding survivors. 

UAV-based SAR missions are quite challenging by nature, with complex objectives and tight constraints. Take for instance the real-world scenario of a missing person in the woods depicted in Fig. \ref{fig:scenario}(b). Specifically, the figure shows a birds-eye-view of the scene from a real-world deployment of a UAV in a SAR mission that the authors participated to assist civil protection officers in a missing person incident. The scene depicts the first responders and their vehicles after the UAV took off and the surrounding rural scenery composed of various forms of vegetation. In the depicted thermal image, lighter color represents higher temperatures. Evidently, trees and high foliage block any thermal dissipation from being detected and thus make it impossible to search for the missing person with this top-down view. Additionally, missing people tend to find cover under tall trees to safeguard themselves from the elements of bad weather which makes detection from above impractical. Although, the use of the UAV in this scenario can provide a first assessment of the situation at hand, the chances of finding the missing person using this tactic are very limited. 

Motivated by the challenges described above, in this work we proposed a UAV-based 3D search-planning framework which can be utilized by first responders during search and rescue missions. More specifically our contributions are the following:

\begin{itemize}
\item We propose an integrated assessment and search planning approach which can be used to aid search-and-rescue missions with an autonomous UAV agent. In the assessment planning stage, a high-level plan is produced which guides the UAV to fly over the area of interest and gather mission critical information. Then in the second stage, i.e., the search planning stage, the generated search plan is further fine-tuned to allow the UAV agent to search the objects of interest in 3D (i.e., across all faces).
\item In order to accomplish the 3D search task a number of artificial rectangular cuboids are generated and placed around the objects of interest that need to be searched. Then, a model predictive control (MPC) algorithm with linear and binary constraints is proposed, for guiding the UAV agent through the generated cuboids. The proposed formulation allows the problem to be solved exactly and efficiently using off-the-shelf solvers.
\item We demonstrate the capabilities and performance of the proposed approach through extensive simulated search scenarios.
\end{itemize}

The rest of the paper is structured as follows. Section~\ref{sec:Related_Work} presents an overview of the related work on UAV-based trajectory planning for SAR missions. Section~\ref{sec:problem} presents the proposed system architecture and Section \ref{sec:system_model} develops the system model. Section \ref{sec:stage1} discusses the details of the assessment planning phase and Section \ref{sec:stage2} presents the proposed 3D search planning phase. Finally, Section \ref{sec:Evaluation} evaluates the proposed system and Section \ref{sec:conclusion} concludes the paper.

\section{Related Work}\label{sec:Related_Work}

Autonomous planning and control are perhaps the two most desired capabilities in mobile robotics. Over the last years a plethora of methods have been proposed from academic and industrial research labs especially for the problem of planning and control of ground robots operating in 2D environments. Although the proposed approaches have reached a significant level of maturity, there are still challenges to be tackled when more complex scenarios are considered i.e., planning and control in 3D with UAVs during search and rescue (SAR) missions. In such scenarios the task of planning and control is considerably more challenging mainly due to the more demanding mission objectives (e.g., searching with a UAV for survivors during natural disasters in three-dimensional environments, scanning around buildings for people in need, etc.). 

UAV-based 3D trajectory planning was investigated in \cite{Bortoff2000,He2008,Tisdale2009,Yang2008}, with the main objective being the search for a collision-free trajectory to the goal region. More specifically, \cite{Bortoff2000} proposes a two-step trajectory planning approach using Voronoi graph search and artificial forces. In \cite{He2008}, the authors investigate the problem of UAV trajectory planning in GPS-denied environments. To handle the inherited uncertainty in these situations the authors extend the Belief Roadmap \cite{Prentice2009} (BRM) planning algorithm with accurate state-estimation based on stochastic-filtering techniques. The trajectory planning problem studied in this work however, is purely kinematic and ignores the UAV dynamics. In \cite{Tisdale2009}, a receding horizon trajectory planning approach is proposed for UAVs and solved using gradient-based methods whereas in \cite{Yang2008} the authors use rapidly-exploring random trees (RTTs) to generate collision-free waypoints in a computationally efficient way. At a second stage the generated waypoints are connected with straight line segments and the resulting path is smoothed out using cubic Bezier curves to create a continuous curvature path which the UAV can execute.
Because of the complex and not flat terrain, the authors in \cite{Colas2013} develop a 3D path planning and execution method based on $\text{D}^*\text{-Lite}$ for search-and-rescue ground robots. In order to reduce the computational complexity of the task, the authors propose to decouple the problem into positioning and orientation planning.

The work in \cite{Berger2015} proposes a multi-agent path-planner for detecting a static target with unknown location during search and rescue missions. The proposed technique is exact and is solved using mixed-integer linear programming (MILP). However, it is based on a 2D discrete representation of the world and it does not considers the system dynamics. 

The authors in \cite{san2018} investigate various discrete path-planning techniques for searching survivors with UAVs during disasters, including artificial potential fields (APF), fuzzy logic and genetic algorithms (GA). The works in \cite{savvas1,savvas2,savvas3,savvas4} develop searching-and-tracking (SAT) approaches for searching an area of interest with multiple UAV agents and tracking multiple targets of interest. These works however are myopic and are based on a 2D representation of the world. 

More recently the work in \cite{Alcantara2019} investigated the problem of search planning during SAR missions using UAVs. The authors formulate the trajectory planning problem as a model predictive control (MPC) problem and they solve it using particle swarm optimization (PSO). This technique used a 2D coordinated kinematic model for the UAVs and the search planning was conducted in two-dimensions. 
Detailed surveys discussing the various trajectory/path planning methods in the literature and their applications can be found in \cite{Yang2014,Radmanesh2018,Yang2016,Souissi2013}.

Compared to the related work the most notable differentiating factors of this work are a) the proposed two-stage search planning architecture which captures the increasing level of complexity of the various phases of a SAR mission and b) the necessity to search in 3D the objects of interest from all views in order to search for survivors.

\section{System Architecture}\label{sec:problem}
\begin{figure*}
	\centering
	\includegraphics[width=\textwidth]{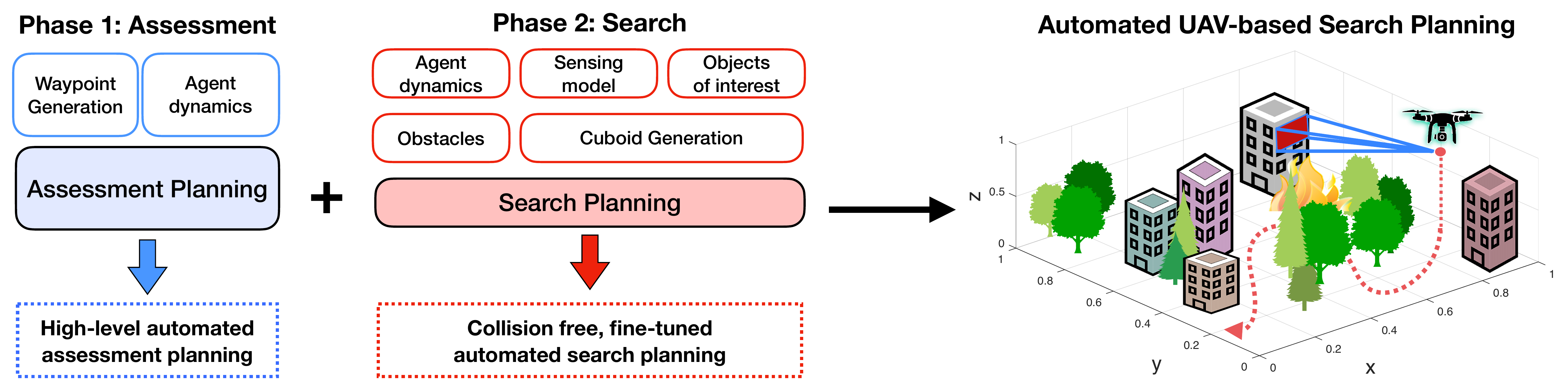}
	\caption{Overview of the proposed search planning system architecture. The proposed framework aims to automate the first two phases (i.e., the assessment phase and the search phase) of a traditional SAR mission with a UAV agent.}	
	\label{fig:arch}
	\vspace{-1mm}
\end{figure*}

SAR missions typically consist of three phases \cite{UNCHR1,UNCHR2} namely a) assessment, b) search and c) rescue. The goal of the assessment phase is the determination of the course of action. This is a critical phase in which the rescue team assesses the damages and the hazards in the vicinity of the affected area in order to prepare and organize the search and rescue mission. On the other hand the goal of the search phase is to conduct efficient, organized and thorough searches in the affected area in order to located survivors as efficiently as possible. Search operations when possible follow optimized search patterns which have been planned ahead in order to increase the efficiency of the search. In addition, the search team is often required to search around and along large structures/buildings, below bridges and under high foliage in order to locate people in need. Finally, during the last phase, people are given medical aid and are transported to safety. 

The proposed approach aims to integrate and automate the first two phases (i.e., the assessment and the search phases) of a traditional SAR mission with a single UAV agent. The proposed approach is illustrated in Fig. \ref{fig:arch}. In the first phase the human operator specifies the critical areas (as waypoints) to be visited in order for the rescue team to assess the incident at hand. In the second phase a more detailed specification is given regarding the mission which allows the UAV agent to search the objects of interest across all faces in 3D.

%

In this paper we assume that a controllable flying agent or a UAV, operates inside a bounded 3D environment which contains a) objects of interest which need to be searched (i.e., searched across all faces) and b) obstacles that need to be avoided. 

Let the set of all objects of interest inside the surveillance region that need to be searched be denoted by $J=\{j_1,j_2,\cdots,j_{|J|}\}$, with the set cardinality $|J|$ denoting their total count. Similarly, we denote the set of all obstacles in the environment by $\Xi=\{\xi_1,\xi_2,\cdots,\xi_{|\Xi|}\}$. We assume that the UAV agent is controllable and  evolves in time according to a discrete-time dynamical model. Additionally, we assume that the agent is equipped with a camera system, with finite field-of-view (FoV), which uses for perceiving its surroundings. 

In the first stage, termed \textit{Assessment planning}, we aim to derive a high-level 3D search plan in order to acquire more information about the objects of interest in the set $J$. This plan is generated in accordance to the agent dynamics and optimizes a simplified mission objective (i.e., minimizing the proximity of the agent with a set of waypoints). In this stage, the 3D objects of interest are approximated by waypoints in 3D, produced in a pre-processing step (i.e., during the waypoint generation step). The generated trajectory of this stage will guide the UAV to fly over the objects of interest by visiting their waypoint approximation, gathering information about the mission such as the dimensions of the objects of interest and the location of the obstacles in the environment. Essentially, the generated UAV trajectory of this phase is used to build a 3D map of the environment \cite{Re,Rf,Rg,Rh}, which in turn is utilized to reconstruct the objects of interest $J$ and obstacles $\Xi$ in the environment as rectangular cuboids for the \textit{Search Planning} stage.

In the second stage, termed \textit{Search Planning}, all objects of interest and obstacles are represented as rectangular cuboids based on the information acquired in the first stage. In this stage, the planning takes place over a smaller area of the surveillance region and produces accurate and fine-tuned plans which take into account the agent dynamics, the environmental constraints (i.e., obstacles), and the agent sensing model (i.e., the specifications of the UAV onboard camera system) and allows the agent to search in 3D all objects of interest while avoiding collisions with the obstacles in the environment. In essence, during this stage we find the optimal UAV control inputs which allow the UAV to search an object of interest across all faces.

\section{System Model} \label{sec:system_model}

\subsection{Agent Dynamical Model}
In this work we assume that the UAV agent evolves in 3D space according to the following discrete-time dynamical model:
\begin{equation} \label{eq:agent_dynamics}
    x_t = \Phi x_{t-1} + \Gamma u_{t-1}
\end{equation}
where $x_t = [\text{x},\dot{\text{x}}]_t^\top \in \mathbb{R}^6$ denotes the agent's state at time $t$ which consists of position $\text{x}_t=[p_x, p_y, p_z]_t \in \mathbb{R}^3$ and velocity $\dot{\text{x}}_t = [\nu_x,\nu_y,\nu_z]_t \in \mathbb{R}^3$ components in 3D cartesian coordinates. The agent can be controlled by applying an amount of force $u \in \mathbb{R}^3$ in each dimension, thus $u_{t} = [f_x, f_y, f_z]_{t}^\top$ denotes the applied force vector at $t$. The matrices $\Phi$ and $\Gamma$ are given by:
\begin{equation}
\Phi = 
\begin{bmatrix}
    \text{I}_3 & \Delta T \cdot \text{I}_3\\
    \text{0}_3 & \phi \cdot \text{I}_3
   \end{bmatrix},~
\Gamma = 
\begin{bmatrix}
    \text{0}_3 \\
     \gamma \cdot \text{I}_3
   \end{bmatrix}
\end{equation}

\noindent where $\Delta T$ is the sampling interval, $\text{I}_3$ is the identity matrix of dimension $3 \times 3$ and $\text{0}_3$ is the zero matrix of dimension $3 \times 3$. The parameters $\phi$ and $\gamma$ are further given by $\phi = (1-\eta)$ and $\gamma = m^{-1} \Delta T$, where $\eta$ is used to model the air resistance and $m$ is the agent mass.

\subsection{Agent Sensing Model}
\begin{figure}
	\centering
	\includegraphics[scale = 0.5]{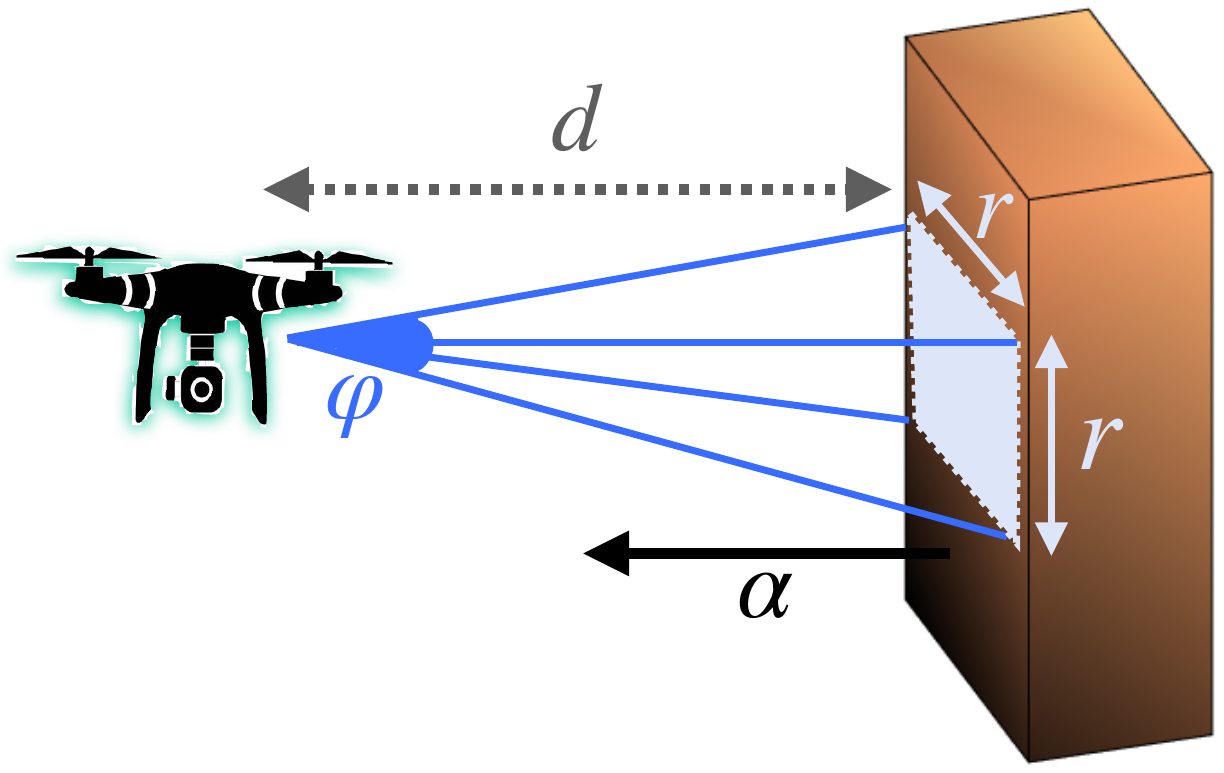}
	\caption{The figure illustrates the agent sensing model i.e., a square region with dimensions $r \times r$. The angle $\varphi$ determines the FoV opening and $\boldsymbol{\alpha}$ is the normal vector to the face of the object. }	
	\label{fig:camera_model}
	\vspace{-0mm}
\end{figure}

The agent is equipped with an onboard camera taking snapshots of the objects of interest in order to search for survivors or people in need. Without loss of generality, we assume in this work that the camera field of view  (FoV) angles in the horizontal and vertical axis are equal \cite{Petrides2017} and thus the projected FoV footprint on a planar surface is square with side length $r$ and given by:
\begin{equation}\label{eq:camera_model}
    r(d) = 2 d  \tan\left(\frac{\varphi}{2}\right)
\end{equation}
where $d$ denotes the distance in meters between the location of agent and the surface of the object that needs to be searched and $\varphi$ is the angle opening of the FoV according to the camera specifications. Thus the area of the FoV footprint at a distance $d$ is $r(d)^{2}$ meters. Before taking a snapshot of the object of interest the agent first aligns its camera with respect to the surface in such a way so that the optical axis of the camera (i.e., the viewing direction) is parallel to the normal vector ($\boldsymbol{\alpha}$) of the surface, as illustrated in Fig. \ref{fig:camera_model}.

In order to search an object of interest the agent needs to take multiple snapshots (according to the size of the FoV as given by Eqn. \eqref{eq:camera_model}) such that the surface area of each face of the object is completely included in the acquired images.

\subsection{Object Representation} \label{ssec:objects}

The objects of interest that need to be searched in 3D and the obstacles inside the surveillance area that need to be avoided are represented in this work as rectangular cuboids of various sizes (referred to as cuboids hereafter). A rectangular cuboid is a convex hexahedron in three dimensional space which exhibits six rectangular faces (i.e., where each pair of adjacent faces meets in a right angle). A point $\mathbf{x}=[x,y,z]^\top \in \mathbb{R}^3$ that belongs to the cuboid $\mathcal{C}$ or resides inside the cuboid satisfies the following system of linear inequalities:
\begin{align}
    a_{11}x + a_{12}y & + a_{13}z  \le b_1 \notag \\
    a_{21}x + a_{22}y & + a_{23}z  \le b_2 \notag \\
    \vdots & \notag \\
    a_{n1}x + a_{n2}y & + a_{n3}z  \le b_n \notag
\end{align} 
where $n=6$ is the total number of faces which compose the cuboid $\mathcal{C}$, $\boldsymbol{\alpha_i} = [a_{i1},a_{i2},a_{i3}]$ is the outward normal vector to the $i_\text{th}$ face of the cuboid and $b_i$ is a constant. In more compact form the expression above can be written as $A \mathbf{x} \le B$ where $A$ is a $n \times 3$ matrix, $\mathbf{x}$ is a $3 \times 1$ column vector and $B$ is a $n \times 1$ column vector. For the rest of the paper, we will denote a rectangular cuboid $\mathcal{C}$ by the matrices $A$ and $B$. That said, in this work we assume that the surveillance region has been 3D mapped \cite{Re,Rf} during the assessment phase and subsequently the objects of interest and obstacles in the environment have been represented as cuboids, as discussed above. Moreover, we should note here that the proposed technique can be applied to any type of object as long as it can be represented as a convex polyhedron.

\section{Assessment Planning} \label{sec:stage1}
In the assessment planning phase the objective is to compute the UAV control inputs which guide the agent to fly over the objects of interest with the ultimate purpose of collecting critical information regarding the mission. We assume that this information includes the dimensions of the objects of interest and obstacles and the faces of the objects of interest that need to be searched. This information is passed to the second stage which is responsible for guiding the UAV agent to search the objects of interest in 3D while avoiding collisions with obstacles in the environment.

In this stage, the objects of interest are approximated by single 3D waypoints and by taking into account the agent dynamics a high-level plan is generated which minimizes the proximity of the agent with the waypoints. Thus the waypoint generation pre-processing step is applied first in which each object of interest $j \in J$ is approximated by a single location in 3D space (i.e., the waypoint $w_j$) and a mixed integer linear program $(P1)$ is derived below to compute the plan which guides the UAV as close as possible to the waypoints.

\begin{algorithm}
\begin{subequations}
\begin{align}
&\textbf{Problem } \texttt{Assessment Planning}: & & & \nonumber\\
& \text{(P1)}~~\underset{u_{0:T-1}}{\min} \sum_{t=1}^{T}\sum_{j=1}^{|J|} \zeta_{t,j} & \label{eq:objective_P1} \\
&\textbf{subject to} ~ t \in \{1,\ldots,T\} \textbf{:}  &\nonumber\\
& x_t = \Phi^t x_0 + \sum_{\tau=0}^{t-1} \Phi^\tau \Gamma u_{t-\tau-1} &  \label{eq:P1_0}& \\
& xw^+_{t,j} - xw^-_{t,j} = d(Hx_t,w_j) & \forall t, j\label{eq:P1_1}&\\
& \sum_{t=1}^T b_{t,j} = 1 & \forall j \label{eq:P1_2}&\\
& \zeta_{t,j}\leq Mb_{t,j}  & \forall  t, j \label{eq:P1_3}&\\
& \zeta_{t,j}\leq xw^+_{t,j} + xw^-_{t,j} & \forall t, j \label{eq:P1_4}&\\
& \zeta_{t,j}\geq xw^+_{t,j}+xw^-_{t,j}-M(1-b_{t,j})& \forall t, j\label{eq:P1_5}&\\
& \{xw^+_{t,j}, xw^-_{t,j},\zeta_{t,j}\}\geq 0 &\forall t, j \label{eq:P1_6}&\\
& b_{t,j}\in\{0,1\} &\forall t, j & 
\end{align}
\end{subequations}
\end{algorithm}

\noindent The program in $(P1)$ finds the optimal control inputs $u_{0:T-1}$ over the planning horizon $T$ which minimizes the $L_1$-norm between the computed UAV 3D trajectory and the waypoints according to the agent dynamics defined in Eqn. \eqref{eq:P1_0}. First, we linearize the $L_1$-norm between the position of the agent $Hx_t$ at a specific time instance and the position of waypoint $w_j$ of object $j$ i.e., $|Hx_t-w_j|$, where $H$ is a matrix which extracts the position coordinates from the agent state $x_t$. 

To do that we define $d(Hx_t,w_j) = Hx_t-w_j$, which according to the constraint in Eqn. \eqref{eq:P1_1}, is assigned to the variable $xw^+_{t,j}$ if it is positive and to the variable $xw^-_{t,j}$ if it is negative. Because, $xw^+_{t,j}$ and $xw^-_{t,j}$ are tied to the decision variable $\zeta_{t,j}$ which is minimized and since $xw^+_{t,j}$, $xw^-_{tj}$ and $\zeta_{t,j}$ are all defined to take positive values (i.e., Eqn. \eqref{eq:P1_6}), the optimizer assigns the value of $|Hx_t-w_j|$ to $\zeta_{t,j}$ via the constraints in Eqn. \eqref{eq:P1_3} - Eqn. \eqref{eq:P1_5} when the binary variable $b_{t,j}$ is activated.

More specifically, the binary variable $b_{t,j}$ indicates the time instance that waypoint $w_j$ is visited by the agent. Thus the decision variable $\zeta_{t,j}$ is assigned with distance values at the time-steps which specific waypoints are visited. The constraint in Eqn. \eqref{eq:P1_1} computes the $L1$-norm between the UAV trajectory and the waypoints for all possible UAV positions in the planning horizon. The constraint in Eqn. \eqref{eq:P1_2} ensures that the number of binary variables that are activated are equal to the total number of waypoints i.e., each waypoint is visited exactly once during the planning horizon. Finally, the set of constraints \eqref{eq:P1_3}-\eqref{eq:P1_5} assign the distance between waypoint $j$ and the UAV position $Hx_t$ at time $t$ to $\zeta_{t,j}$, if $b_{t,j}=1$. Observe, that when $b_{t,j}=0$ the value of $\zeta_{t,j}$ is driven to zero. Thus, the minimization of Eqn. \eqref{eq:objective_P1} forces the UAV agent to approach as close as possible to the waypoints, one waypoint at a time, while enforcing the constraint of its dynamical model. 

The produced 3D trajectory guides the agent to pass from all waypoints thus acquiring information regarding the objects of interest i.e., their dimensions and other critical information about the mission such as the location of obstacles which are used in stage two below.

\section{Search Planning} \label{sec:stage2}

In the second phase, termed \textit{Search Planning}, the framework can be used to compute fine-tuned and collision-free search plans inside a smaller area of the surveillance region. In particular in this stage the generated search plans allow the agent to search the objects of interest in 3D (i.e., for each object of interest all their faces are searched for survivors according to the camera specifications). We assume that the information collected during the first stage allows us to represent each object of interest $j$ that needs to be searched and the obstacles in the environment as rectangular cuboids i.e., Sec. \ref{ssec:objects}. Thus for this phase we assume that the UAV has obtained a map of the environment which contains the locations of the objects of interest as well as the locations of the obstacles in the environment. In this map all objects of interest and obstacles have been represented as rectangular cuboids of appropriate dimensions.

Before discussing the details of the proposed search planning technique we first describe the pre-processing step (i.e., the cuboid generation step) which takes place before the main trajectory planning process.

\begin{figure}
	\centering
	\includegraphics[width=\columnwidth]{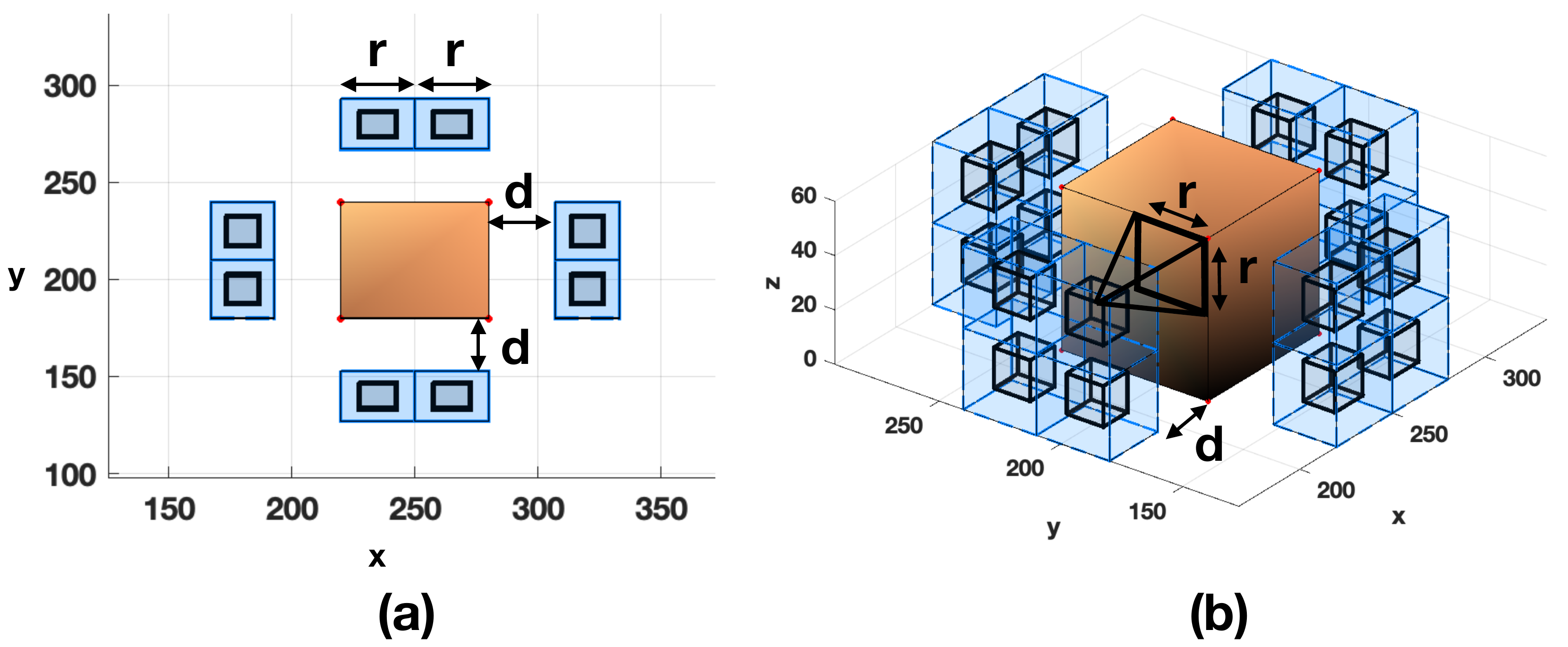}
	\caption{The figure illustrates the cuboid-generation pre-processing step which allows the agent to search all faces of an object of interest by passing through the generated cuboids.}	
	\label{fig:tiles}
	\vspace{-0mm}
\end{figure}  

\subsection{Cuboid Generation} 
The main idea here is to generate artificial cuboids all around the object of interest and then guide the UAV agent through these cuboids in order to search the object in 3D from all faces, i.e., scan the total surface area of the object with the UAV's camera system. The generated cuboids are placed at specific distances from the object of interest according to the specifications of the UAV camera system Eqn. (\ref{eq:camera_model}). More specifically, we assume that we know a-priori the distance $d$ that the UAV must maintain with a particular object of interest during searching. Given $d$ and the camera FoV angle $\varphi$, the size of the UAV FoV footprint can be determined by Eqn. (\ref{eq:camera_model}) as a square with length $r$. Subsequently, the area of each face of the object of interest that needs to be searched is decomposed into several $r \times r$ cells and for each of those cells a cuboid is generated at distance $d$, so that when the agent passes from within a cuboid the corresponding area of the face is searched. That said, in the cuboid generation preprocessing step, for each face $f_i, i \in [1,\ldots,6]$ of the object of interest $j$ we generate artificial cuboids $\mathcal{C}^j_{i,n}, n \in [1,\ldots,N^j_i]$ at distance $d$ where $N^j_i$ is the total number of cuboids that are necessary to cover the whole area of face $f_i$ of object of interest $j$.

This is illustrated in Fig. \ref{fig:tiles}, where 4 faces of the object of interest need to be searched and thus each of the 4 faces has been decomposed into several cells marked with light blue color. For each cell a cuboid shown in black color is generated to allow the agent to pass through it and search the corresponding part (area) of the face. 
More specifically, in Fig. \ref{fig:tiles}, the object of interest (with brown color) is represented by a rectangular cuboid with dimensions width=$60$m, length=$60$m and height=$60$m. We assume that this information was acquired during the Assessment planning phase. The UAV agent must maintain a distance of $d=27$m from each of the object's faces, and the FoV angle is $\varphi=60$deg. The camera FoV area in this case is approximately $30 \times 30 ~\text{m}^2$ according to Eqn. (\ref{eq:camera_model}). That said, each of the objects's faces is decomposed into cells of size  $30 \times 30 ~\text{m}^2$ creating a grid of 4 cells. This is depicted in Fig. \ref{fig:tiles} with the light blue 3D boxes in front of each face. Then for each cell a cuboid is generated at distance $d$ from the object, shown in black color in Fig. \ref{fig:tiles}. When the agent resides within a cuboid their FoV projection captures an area of size $30 \times 30 ~\text{m}^2$ on the face of the object. By guiding the agent through all 4 cuboids we allow the agent to cover the total surface area of a face with its camera system. Consequently, we are interested in guiding the agent via all the generated cuboids in order to search the object of interest across all faces.

\subsection{3D Search Planning}
Once the cuboid generation pre-processing step is completed, we formulate the problem of searching the object of interest in 3D, as a rolling-horizon model predictive control (MPC) problem with linear and binary constraints and we solve it using  off-the-shelf mixed-integer programming (MIP) solvers. Specifically, in the proposed approach at each sampling time $t$ the UAV control actions  $u_{t|t},\ldots,u_{t+T-1|t}$ are obtained over a rolling-horizon of length $T$ time-steps by solving a finite horizon open-loop optimal control problem using the current state of the UAV $x_{t|t}$ as the initial state. The first control action $u_{t|t}$ in the sequence is then applied to the UAV and the optimization is repeated for the next sampling time. In this problem we consider 2 different types of constraints: a) linear constraints which govern the UAV dynamical model and b) binary constraints for the search task, obstacle avoidance and duplication of effort. Let us assume, that the cuboid generation pre-processing step has created a total of $N$ cuboids $\mathcal{C}_n, n \in [1,\ldots,N]$ (across all faces) around a single object of interest $j$. For notational clarity we assume that for every face $i$ of the object of interest we generate the same number of cuboids and we drop the index $j$ of the object of interest, thus $\mathcal{C}^j_{i,n}$ becomes $\mathcal{C}_{n}$ for a single object of interest.

We associate each cuboid $\mathcal{C}_n$ that needs to be visited with a binary variable $y_{n}$ which indicates whether this cuboid has been marked to be visited at some future time step $(t+\tau+1|t), \tau \in \{0,\ldots,T-1\}$ i.e., $y_{n} = 1, ~\text{iff} ~Hx_{t+\tau+1|t} \in \mathcal{C}_n$. Here $H$ is a matrix which extracts the position coordinates from the agent's state. That said, the 3D search task objective becomes the maximization of the cuboids to be visited over the planning horizon i.e.,
\begin{equation} \label{eq:visiting}
    \max \sum_n y_{n}
\end{equation}

For infinite planning horizon problems Eqn. \eqref{eq:visiting} can be used to devise a search plan that will visit all the unvisited cuboids. However, to apply Eqn. \eqref{eq:visiting} in a rolling finite horizon fashion we need to keep track of the visited cuboids in order to avoid visiting the same cuboids over and over. To achieve this, the UAV agent maintains a map $V$ which is used to keep track of all visited cuboids as shown below:
\begin{equation}
 V(n) = 
  \begin{cases} 
   1 & \text{,~if } Hx_{0:t} \in \mathcal{C}_n \\
   0 & \text{,~o.w} \\
  \end{cases}
\end{equation}

\noindent where $Hx_{0:t}$ denotes the agent's locations up to time $t$. That said, the complete mathematical formulation of the 3D search planning problem tackled in this work is shown in (P2) below:

\subsubsection{Mission Objective}
In (P2) we are interested in finding the optimal UAV control actions $u_{t|t},\ldots,u_{t+T-1|t}$ over the rolling horizon of length $T$ that minimizes the mission objective $h(\mathbf{x},\mathbf{u},\mathbf{y})$ shown in Eqn. \eqref{eq:mission_objective}. The mission objective is a function of the agent's future state $\mathbf{x}$, control inputs $\mathbf{u}$ and the binary variables $\mathbf{y}$ which indicate whether a cuboid has been planned to be visited in the future. More specifically, the mission objective is given by:

\begin{align} \label{eq:mission_objective}
    &h(\mathbf{x},\mathbf{u},\mathbf{y}) =a \|Hx_{t+\tau^\star+1|t}-x^\star\|^2_2 ~~+\\ \notag &
    ~~~~~~~~b\sum_{\tau=1}^{T-1} \|u_{t+\tau|t}-u_{t+\tau-1|t}\|^2_2 ~-~ c\sum_{n=1}^{N} y_{n}  &
\end{align}

\noindent where $\tau^\star \in \{0,\dots,T-1\}$, $x^\star$ is the centroid of the nearest (with respect to the agent's current location) unvisited cuboid and $(a, b, c)$ are weights associated with the different mission objectives. In particular, the first term i.e., $\|Hx_{t+\tau^\star+1|t}~-~x^\star\|^2_2$ guides the UAV agent towards the nearest unvisited cuboid by minimizing the squared euclidian distance of the future agent location $Hx_{t+\tau^\star+1|t}$  with the centroid of the nearest unvisited cuboid $x^\star$. Moreover, the second term i.e., $\sum_{\tau=0}^{T-1} \|u_{t+\tau+1|t}-u_{t+\tau|t}\|^2_2$ is used to minimize the deviations between consecutive control inputs, thus leading to smoother UAV trajectories. Finally, the last term in Eqn. \eqref{eq:mission_objective} is used for maximizing the number of cuboids to be visited over the planning horizon, as discussed previously.

\subsubsection{Mission Constraints}
We can now describe the mission constraints which lead to the desired 3D search planning behavior. The constraints in Eqn. \eqref{eq:P2_1} - \eqref{eq:P2_8} are due to the agent dynamical model as described in Section \ref{sec:system_model}, assuming a known initial state $x_{1|1} = x_0$. As we have already mentioned, the cuboid generation pre-processing step has created a total of $N$ cuboids $\mathcal{C}_i, i \in [1,\ldots,N]$ for the object of interest $j$.

The constraints in Eqn. \eqref{eq:P2_2}-\eqref{eq:P2_4} are due to the 3D search task. Specifically, the constraint in \eqref{eq:P2_2} uses $\tau \times N \times L$ binary variables $b_{\tau,n,l}$, for $\tau \in \{0,\ldots,T-1\}$, $n \in \{1,\ldots,N\}$ and $l \in \{1,\ldots,L\}$ to determine whether the agent position $Hx_{t+\tau+1|t}$ resides inside the negative or positive half-space defined by the plane which contains the $l_\text{th}$ face of the $n_\text{th}$ cuboid. The matrix $A_n$ of size $(L \times 3)$ and the vector $B_n$ of size $(L \times 1)$ represent the linear system of inequalities $A_n x \le B_n$ for which a 3D point $x$ must satisfy in order to reside inside the cuboid $\mathcal{C}_n$ i.e., Sec. \ref{ssec:objects}. Thus, the inequality $A_{n,l} Hx_{t+\tau+1|t} \le B_{n,l}$ becomes true when the UAV location resides inside the negative half-space created by the plane containing the $l_\text{th}$ face of the $n_\text{th}$ cuboid. When this happens the corresponding binary variable $b_{\tau,n,l}$ becomes 1 to satisfy the constraint. Otherwise $b_{\tau,n,l}=0$ and the constraint in Eqn. \eqref{eq:P2_2} is valid with the addition of a large positive constant (i.e., big-M) $M$ as shown.

\begin{algorithm}
\begin{subequations}
\begin{align}
&\textbf{Problem } \texttt{Search Planning}: & & & \nonumber\\
& \text{(P2)}~~ \underset{u_{t|t},\ldots,u_{t+T-1|t}}{\min} h(\mathbf{x},\mathbf{u},\mathbf{y}) &\label{eq:objective_P2} \\
&\textbf{subject to} ~ \tau \in \{0,\ldots,T-1\} \textbf{:}  &\nonumber\\
& x_{t+\tau+1|t} = \Phi x_{t+\tau|t} + \Gamma u_{t+\tau|t} & \forall \tau \label{eq:P2_1}\\
& x_{t|t} = x_{t|t-1} & &\label{eq:P2_8}\\
& A_{n,l} H x_{t+\tau+1|t} + (M-B_{n,l}) b_{\tau,n,l} \le M & \forall \tau,n,l \label{eq:P2_2}\\
& L \tilde{b}_{\tau,n} - \sum_{l=1}^L b_{\tau,n,l} \le 0 & \forall \tau,n \label{eq:P2_3}\\
& \hat{b}_n \le \sum_\tau \tilde{b}_{\tau,n}   & \forall n \label{eq:P2_11}\\
& y_{n} \le \hat{b}_{n} + V(n) & \forall \tau,n \label{eq:P2_4}\\
& A_{\psi,l} H x_{t+\tau+1|t} > B_{\psi,l} - M z_{\tau,\psi,l} & \forall \tau,\psi, l \label{eq:P2_5}\\
& \sum_{l=1}^L z_{\tau,\psi,l} \le L-1 & \forall \tau,\psi \label{eq:P2_6}\\
& b_{\tau,n,l},\tilde{b}_{\tau,n},\hat{b}_{n},y_{n},z_{\tau,\psi,l} \in \{0,1\} & \forall \tau,n,\psi,l \label{eq:P2_7}\\
& |\dot{\text{x}}_{t+\tau+1|t}| \le v_\text{max} & \forall \tau \label{eq:P2_9}\\
& |u_{t+\tau+1|t}| \le u_\text{max} & \forall \tau \label{eq:P2_10}
\end{align}
\vspace{-0mm}
\end{subequations}
\end{algorithm}

Subsequently, the binary variable $\tilde{b}_{\tau,n}$ in Eqn. \eqref{eq:P2_3} is activated when the system of linear inequalities is satisfied i.e., $A_{n,l} Hx_{t+\tau+1|t} \le B_{n,l}, \forall l$ which indicates that the agent location $Hx_{t+\tau+1|t}$ resides inside the cuboid $n$ at time $t+\tau+1$.
The constraint in Eqn. \eqref{eq:P2_11} makes sure that the agent is not rewarded by planning trajectories which visit the same cuboid more than once in the future. 

Finally, the constraint in Eqn. \eqref{eq:P2_4} makes sure that the agent is not rewarded for visiting cuboids that have been visited in the past.
 
Lastly, the constraints (\ref{eq:P2_5}) - (\ref{eq:P2_6}) are the collision avoidance constraints with the object of interest and with obstacles in the environment. The UAV agent avoids a collision with an object when:
\begin{equation}
    Hx_{t+\tau+1|t} \notin \mathcal{C}_\psi,~\forall \psi \in \Psi, \tau \in \{0,\ldots,T-1\}
\end{equation}

\noindent where $\mathcal{C}_\psi$ denotes the obstacle $\psi \in  \Psi$ to be avoided which is represented as a convex polyhedron e.g., a rectangular cuboid. Suppose that all objects $\psi$ to be avoided contain $L$ faces, then a collision is avoided at time $\tau$ with object $\psi$ if $\exists l \in \{1,\ldots,L\}: A_{\psi,l} H x_{t+\tau+1|t} > B_{\psi,l}$.

The constraint in Eqn. \eqref{eq:P2_5} uses $T \times |\Psi| \times L$ binary variables $z_{t, \psi, l}$ to check if the constraint $A_{\psi,l} H x_{t+\tau+1|t} > B_{\psi,l}$ is violated where $M$ is a big positive constant. Then, constraint \eqref{eq:P2_6} counts the number of times $z_{t, \psi, l}$ is activated and makes sure that this number is less or equal than $L-1$. Equivalently, agent controls are chosen so that the linear system of inequalities $A_{\psi} H x_{t+\tau+1|t} < B_{\psi}$ is not satisfied. The constraint in Eqn. \eqref{eq:P2_7} declares the binary variables of the problem and finally the constraints in Eqn. \eqref{eq:P2_9} and Eqn. \eqref{eq:P2_10} define the agent's maximum speed and maximum control input respectively. 

Finally, we should mention that the above search planning formulation can be easily extended for multiple objects of interest. This can be achieved by expanding the binary variables by one dimension in order to distinguish the cuboids of different objects of interest. For instance for objects of interest $j \in J$ the binary variable $b_{\tau,n,l}$ in Eqn. \eqref{eq:P2_2} becomes $b^j_{\tau,n,l}$, the binary variable $\tilde{b}_{\tau,n}$ in Eqn. \eqref{eq:P2_3} becomes $\tilde{b}^j_{\tau,n}$ and so forth.

\begin{figure}
	\centering
	\includegraphics[width=\columnwidth]{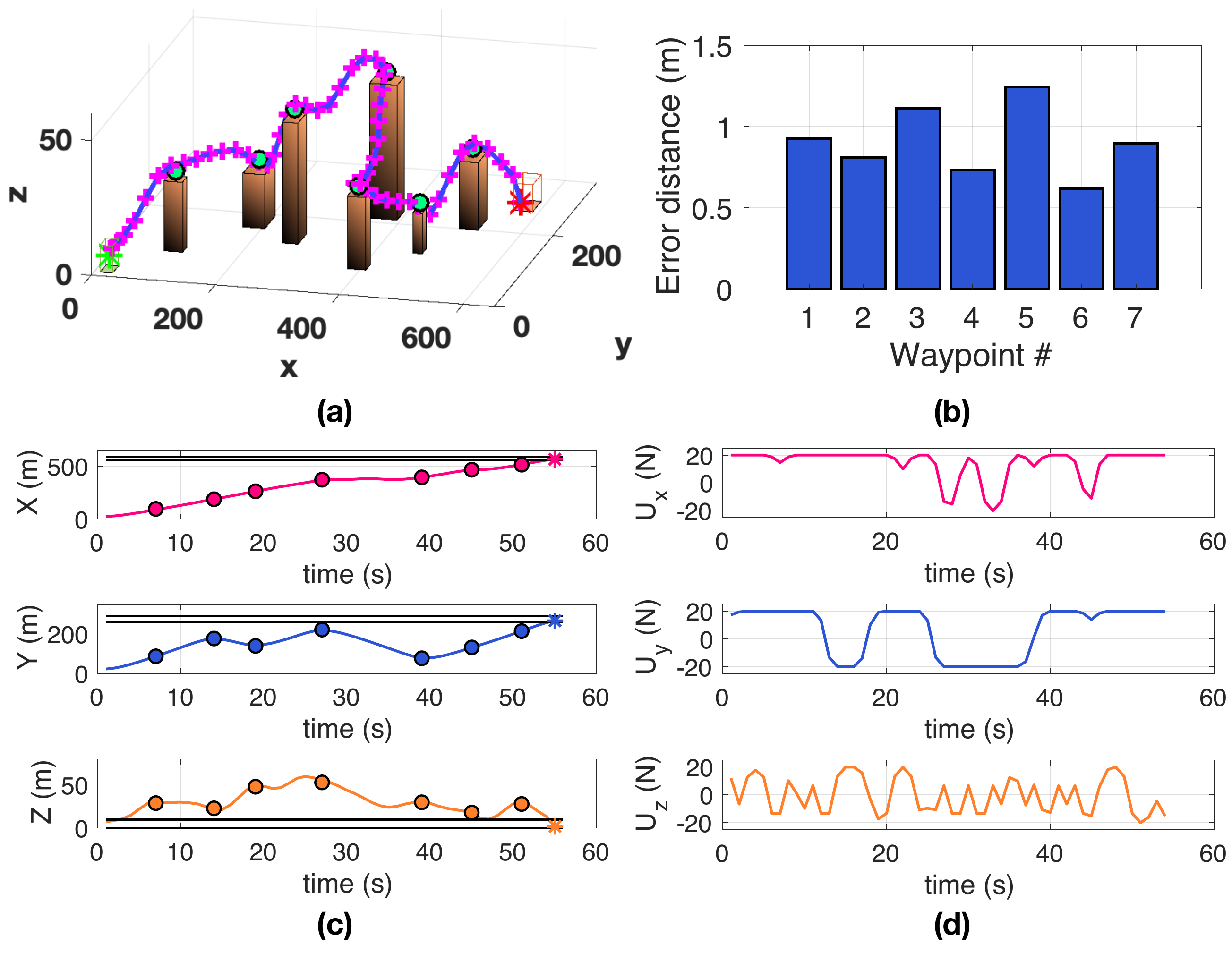}
	\caption{An illustrative example of the Assessment Planning phase}	
	\label{fig:results1}
	\vspace{0mm}
\end{figure}

\section{Evaluation} \label{sec:Evaluation}
\subsection{Experimental Setup}
To evaluate the performance of the proposed two-stage search planning approach we have conducted several synthetic experiments and simulations with varying number of waypoints, obstacles and objects of interest. In each test we evaluate the proposed approach either qualitatively or quantitatively and we discuss its strengths and weaknesses. The experimental evaluation is divided into two parts. First, we investigate the performance of the \textit{Assessment Planning} phase as discussed in Sec. \ref{sec:stage1} and then we conduct the experimental evaluation of the \textit{Search Planning} phase i.e., Sec. \ref{sec:stage2}.

The experimental setup used for the evaluation of the proposed system is as follows: The agent dynamics are expressed by Eqn. (\ref{eq:agent_dynamics}) with $\Delta T = 1$s, agent mass $m=3.35$kg and $\eta = 0.2$. The maximum applied control input $u_\text{max}$ is 20N or (kg.m/$\text{s}^2$), the maximum agent velocity  $v_\text{max}$ is  15m/s and the agent acceleration can reach 6m/$\text{s}^2$. Moreover, we assume that the agent FoV angle $\phi$ is 60deg.

\subsection{Results}
\subsubsection{\textbf{Assessment Planning}}

\begin{figure}
	\centering
	\includegraphics[width=\columnwidth]{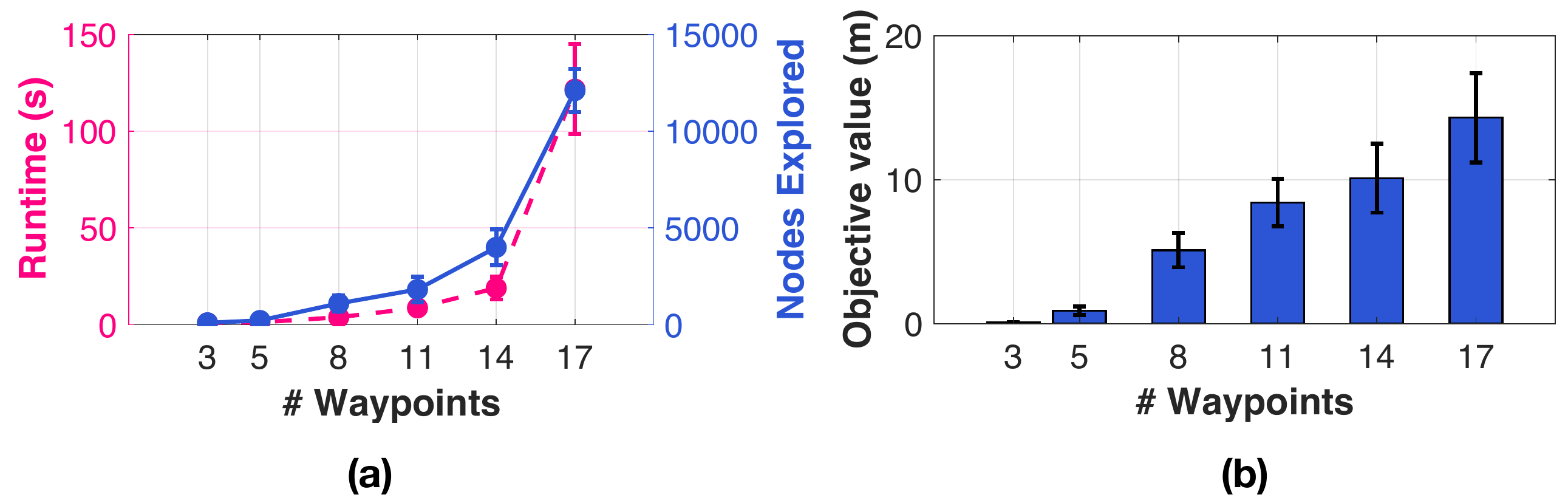}
	\caption{The figure illustrates how the number of waypoints affects the performance of the proposed technique.}	
	\label{fig:results2}
	\vspace{0mm}
\end{figure}

We begin our evaluation with a 3D simulated scenario for the assessment planning phase. In this scenario we deploy 7 objects of interest of various sizes throughout the surveillance area of dimension 600m $\times$ 300m $\times$ 70m. The objects of interest are approximated in this step as 3D waypoints shown as green circles in Fig. \ref{fig:results1}a. The planning horizon $T$ is set to 55 time-steps and the mixed-integer linear program in (P1) is applied to obtain the results shown in Fig. \ref{fig:results1}. The objective here is to devise a plan that will guide the UAV agent to visit all waypoints within the specified time horizon while satisfying the agent dynamical constraints. The agent start position is marked with a green asterisk and the agent final position is marked with a red asterisk as shown in the figure. Figure. \ref{fig:results1}a shows the generated UAV trajectory that visits all the specified waypoints. Figure \ref{fig:results1}b shows the $L1$-distance of the planned trajectory with each of the 7 waypoints and finally Fig. \ref{fig:results1}c and Fig. \ref{fig:results1}d show in more detail the $(x, y, z)$ coordinates of obtained trajectory and the applied control inputs respectively.

\begin{figure*}
	\centering
	\includegraphics[width=\textwidth]{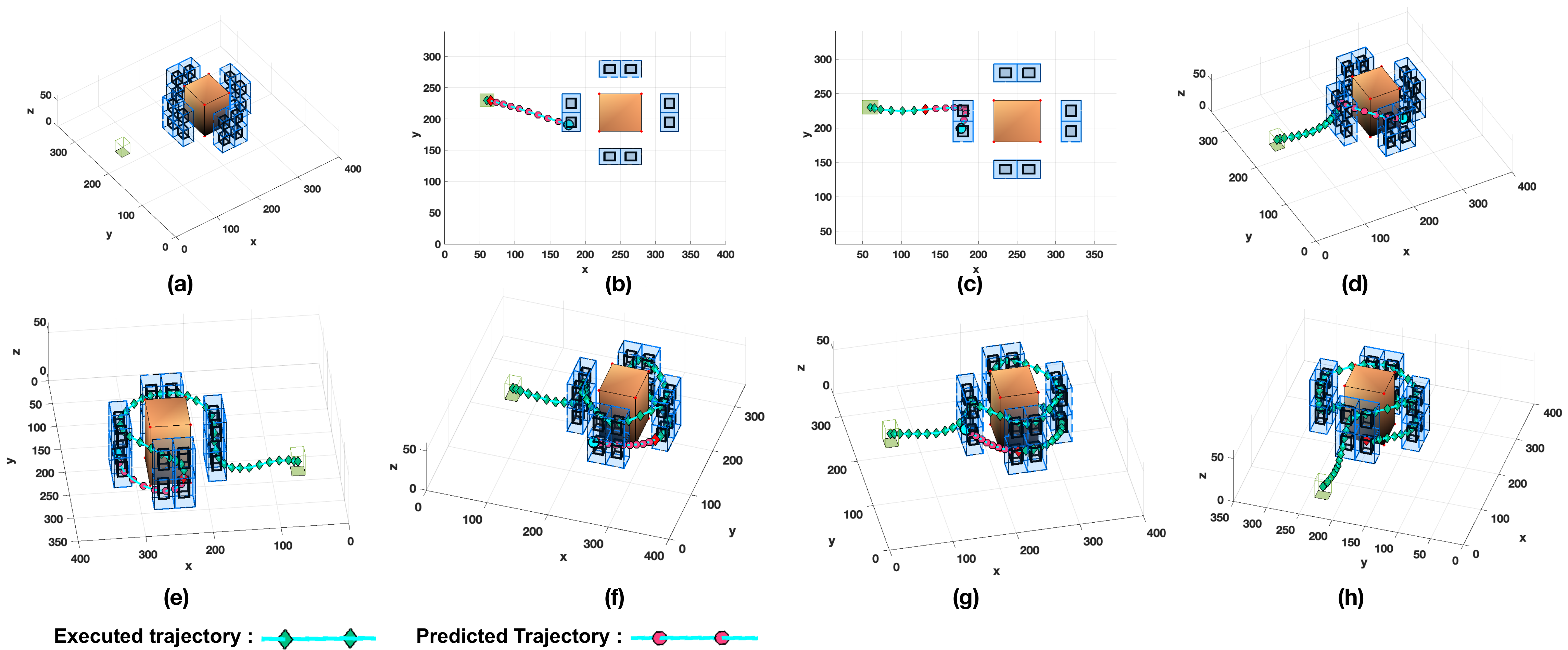}
	\caption{Simulated scenario demonstrating the \textit{Search planning phase} for one object of interest as discussed in Sec. \ref{sec:stage2}.}
	\label{fig:fig5}
	\vspace{0mm}
\end{figure*}

The next experiment aims to investigate how the number of waypoints affect the performance of the proposed assessment planning technique. For this experiment we have conducted 20 Monte Carlo trials where we uniformly generate random waypoints inside a surveillance area of dimensions 200m $\times$ 200m $\times$ 50m and will let our system to run with a fixed time-horizon of 50 time-steps, measuring the runtime, the number of explored nodes during the MIP branch-and-bound optimization and the value of the objective function at the end of the optimization. Figure \ref{fig:results2}a shows the average runtime and the average number of explored nodes as a function of the number of waypoints. As we can observe the average runtime increases as the number of waypoints increases. Additionally, we can observe that the complexity of the problem increases with the number of waypoints. This is also shown by the number of nodes that have been explored until a solution is found. As the number of binary variables increases in a mixed-integer program the produced search tree that is needed to be explored during the branch-and-bound optimization procedure also increases in size and as a result more nodes are needed to be explored until a solution is found. Finally, Fig. \ref{fig:results2}b shows the value of the objective function at the end of the optimization. Essentially, we would like to drive this value close to zero, as we would like to bring the generated search plan as close as possible to the waypoints. However, this is not always possible especially as the number of waypoints increases. This is attributed to the overall complexity of the problem i.e., agent dynamics, the placement of the waypoints, the length of time horizon, etc.

\subsubsection{\textbf{Search Planning}}
The next set of experiments aims to demonstrate the performance of the proposed search planning technique as discussed in Sec. \ref{sec:stage2}. The objective here is to produce fine-tuned and accurate search plans which search the object of interest in 3D across all faces while avoiding collisions with the surrounding obstacles.

The first experiment aims to demonstrate the 3D search task for an object of interest which is represented by a cuboid with size 60m $\times$ 60m $\times$ 60m. In this experiment, the agent is tasked to conduct search planning around the object (i.e., covering four of its faces) at a distance of no less than $d=$27m from the object of interest. At this distance the camera FoV footprint has size 30m $\times$ 30m and thus for each of the object's faces 4 cuboids are generated to capture the total surface area of the face. In total 16 cuboids are generated around the object of interest as illustrated in Fig. \ref{fig:fig5}(a). The agent's initial state is $X_0 = [60,230,10,0,0,0]^\top$, indicated by the green square in Fig. \ref{fig:fig5}(a), and the weights of the objective function in Eqn. \eqref{eq:mission_objective} are set to $(a,b,c) = (0.3, 0.001, 1.5)$. The planning horizon for this experiment is set to 10 time-steps.

In Fig. \ref{fig:fig5}(a), the object of interest to be searched is colored orange and the generated cuboids that need to be visited by the agent are marked in black. Finally, the executed UAV trajectory is marked with green diamonds and the future UAV trajectory over the planning horizon is marked with red circles.

As shown, in Fig. \ref{fig:fig5} the agent passes from all 16 cuboids and thus manages to search all 4 faces of this object. More specifically, Fig. \ref{fig:fig5}(b) - Fig. \ref{fig:fig5}(h) show the agent executed and planned trajectory at time-steps 1, 6, 13, 38, 50, 57 and 63 respectively. As we can observe in each planning horizon the agent tries to maximize the number of cuboids that are visited while at the same time minimizes its distance to the nearest unvisited cuboid according to the mission objective in Eqn. \eqref{eq:mission_objective}. For this scenario, the search planning problem in (P2) requires $10 \times 16 \times 6$ binary variables for $b_{\tau,n,l}$, $10 \times 16$ binary variables for $\tilde{b}_{\tau,n}$, 16 binary variables for $\hat{b}_{n}$ and finally $16$ binary variables for $y_{n}$. Thus, in total we need 1152 binary variables to model the functionality of this problem in an obstacle free environment.

In particular, the main factor that drives the computational complexity is the number of binary variables which are required by (P2). As the number of binary variables increases the search-space that is needed to be explored during the optimization process increases in size and as a result more nodes are needed to be explored until a feasible solution is found. For a problem with $|J|$ objects of interest, $N$ cuboids per object of interest, $|\Psi|$ obstacles and planning horizon of length $T$, the number of binary variables needed by (P2) is equal to $2N|J| + T \left[N |J| (L+1) + |\Psi|L \right]$ (assuming each polyhedron in the environment has $L$ faces) which drives the main computational complexity.

\begin{figure}
	\centering
	\includegraphics[width=\columnwidth]{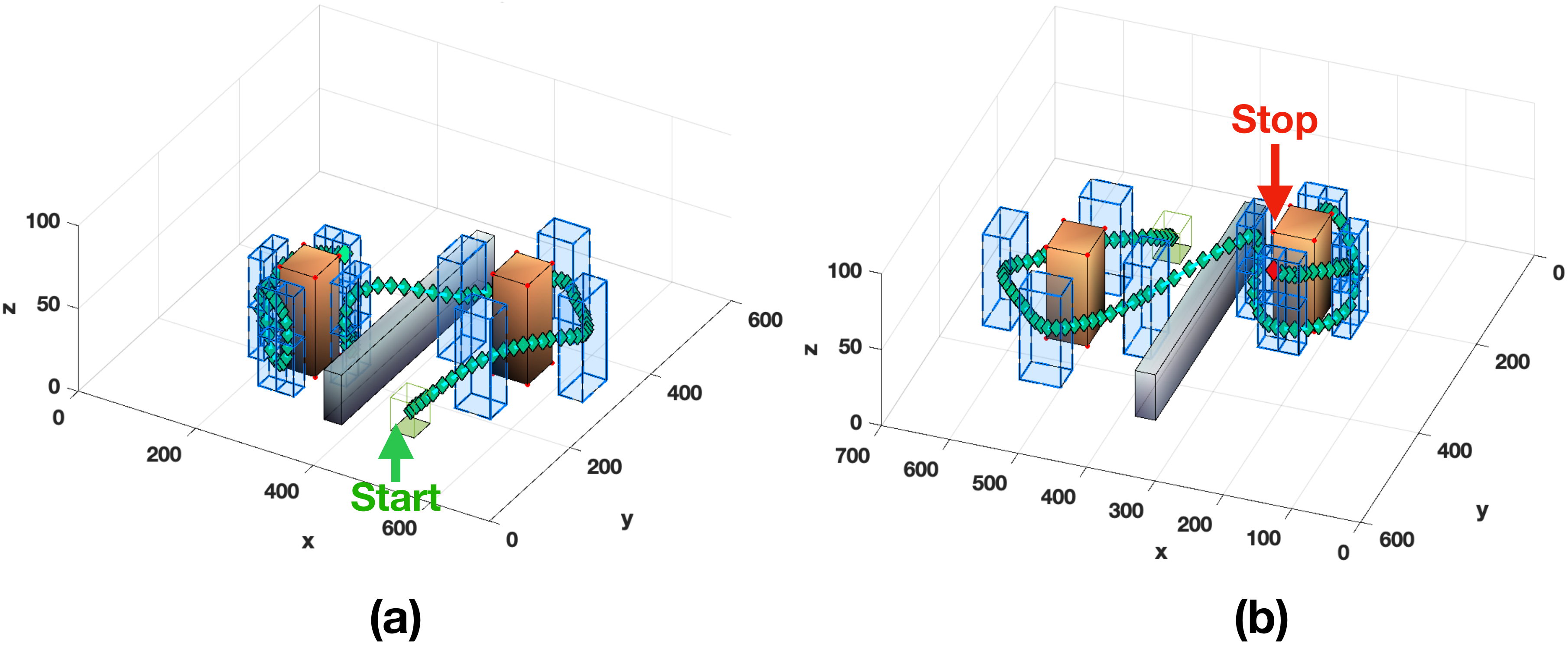}
	\caption{Search Planning for multiple objects of interest in the presence of obstacles.}	
	\label{fig:fig6}
	\vspace{0mm}
\end{figure}

Finally, Fig. \ref{fig:fig6} shows the proposed 3D search planning approach for multiple objects of interest in the presence of obstacles. In this scenario, two objects of interest of size 60m $\times$ 60m $\times$ 60m each are placed inside the surveillance area. The UAV agent is instructed to search the first object from a distance of $d_1=53$m and the second object from a distance of $d_2=27$m. This results in FoV sizes of 60m $\times$ 60m and 30m $\times$ 30m for $d_1$ and $d_2$ respectively. Additionally, the two objects of interest are separated by an obstacle of dimensions 30m $\times$ 380m $\times$ 63m as shown in Fig. \ref{fig:fig6}. The UAV initial state is $X_0 = [460,140,10,0,0,0]^\top$, the planning horizon $T$ is 12 time-steps and and the weights of the objective function in Eqn. \eqref{eq:mission_objective} are set to $(a,b,c) = (0.3, 0.001, 0.001)$. According to the agent FoV size for this scenario each face of the first object was decomposed into one cell (illustrated by the light blue 3d boxes) whereas each face of the second object was decomposed into 4 cells. The cuboid generation pre-processing step generates 4 cuboids (1 cuboid per face) for the first object and 16 cuboids (4 cuboids per face) for the second object. As we can observe, the proposed approach guides the UAV agent to search both objects starting from the first object of interest and moving to the second  while avoiding the obstacle in its path. The final agent location is shown in Fig. \ref{fig:fig6}(b) with a red diamond. Finally we should mention that, in the examples above we have shown the search planning task on the lateral faces of the objects of interest. This is done purely for presentation purposes and for visual clarity. The orientation of the face in 3D space is irrelevant in the proposed approach i.e., a face on the horizontal plane (e.g., the face on the top of the object of interest) can be handled the same way.

\section{Conclusion} \label{sec:conclusion}

In this work we have proposed a novel approach, for integrating and automating the first two phases of a traditional search and rescue mission, with an autonomous UAV agent. In the first stage, namely assessment planning a high-level plan is produced which allows the UAV to fly over the area of interest and gather mission critical information. Then in the second stage i.e., search planning, the generated plan is further fine-tuned to allow the UAV agent to search the objects of interest in 3D (i.e., across all faces) while avoiding collisions with the obstacles in its path. The performance of the proposed approach has been demonstrated through extensive simulation analysis.  
Future work, will investigate the real-world implementation of the proposed system, and its integration into our existing multi-drone tasking platform. Additional future directions include the extension of this work to multiple UAV agents, and the investigation of search planning techniques in noisy and uncertain environments. 

\section*{Acknowledgments}
This work is supported by the European Union's Horizon 2020 research and innovation programme under grant agreement No 739551 (KIOS CoE), by the European Union Civil Protection Call for proposals UCPM-2019-PP-AG grant agreement No 873240 (AIDERS) and from the Republic of Cyprus through the Directorate General for European Programmes, Coordination and Development.

\flushbottom
\balance

\bibliographystyle{IEEEtran}
\bibliography{main} 

\end{document}